%% file: main.tex
\documentclass[10pt,twocolumn,letterpaper]{article}

\usepackage{cvpr}              

\usepackage{graphicx}
\usepackage{caption}
\usepackage[table]{xcolor}
\usepackage{booktabs}
\usepackage{multirow}
\usepackage[table]{xcolor}
\usepackage{colortbl}
\usepackage{array}
\usepackage{tabularx}
\definecolor{accentblue}{RGB}{0,70,140}
\usepackage{pifont}
\usepackage[T1]{fontenc}
\usepackage[utf8]{inputenc}
\usepackage{lmodern}

\input{preamble}
\definecolor{cvprblue}{rgb}{0.21,0.49,0.74}
\usepackage[pagebackref,breaklinks,colorlinks,allcolors=cvprblue]{hyperref}

\NewDocumentCommand{\shilong}
{ mO{} }{\textcolor{red}{\textsuperscript{\textit{Shilong}}\textsf{\textbf{\small[#1]}}}}

\def\paperID{14533} 
\def\confName{CVPR}
\def\confYear{2026}

\title{Chain-of-Ground: Improving GUI Grounding via Iterative Reasoning and Reference Feedback}

\author{
Aiden Yiliu Li\textsuperscript{1,4}\thanks{Work was done during an internship at Princeton AI{$^2$} Lab}  \quad
Bizhi Yu\textsuperscript{2} \quad
Daoan Lei\textsuperscript{2} \quad
Tianhe Ren\textsuperscript{3}\thanks{Corresponding Authors} \quad
Shilong Liu\textsuperscript{4{$\dagger$}} \\
\textsuperscript{1}University College London \quad
\textsuperscript{2}Chico Future AI Lab \quad\\
\textsuperscript{3}The University of Hong Kong \quad
\textsuperscript{4}Princeton University
\\[0.6em]
{\tt\small yiliu.li@outlook.com}\quad {\tt\small slongliu86@gmail.com} 
}

\begin{document}
\maketitle
\input{sec/0_abstract}

\input{sec/1_intro}
\input{sec/2_related_works}
\input{sec/3_methods}
\input{sec/4_tpanel-ui}
\input{sec/5_eval}
\input{sec/6_conclusion}
{
    \small
    \bibliographystyle{ieeenat_fullname}
    \bibliography{main}
}

\end{document}

%% file: sec/0_abstract.tex
\begin{abstract}
GUI grounding aims to align natural-language instructions with precise regions in complex user interfaces (UIs).
While advanced MLLMs have demonstrated strong capabilities in visual GUI grounding, they still struggle with small or visually similar targets, and ambiguity in real-world layouts.
We argue that these limitations stem not only from the models’ inherent grounding capacity, but also from an overlooked underutilization of their existing reasoning potential.
To address this, we present \textbf{Chain-of-Ground (CoG)}, a training-free multi-step grounding framework that leverages MLLMs for iterative visual reasoning and refinement.
Instead of relying on direct prediction, Chain-of-Ground enables the model to progressively reflect and adjust its hypotheses, achieving more accurate and interpretable localization.
Our approach establishes a new state of the art on the ScreenSpot-Pro benchmark with \textbf{68.4\%} accuracy, surpassing the previous best by \textbf{4.8\%}.
To evaluate real-world generalization, we introduce \textbf{TPanel-UI}, a dataset of 420 labeled industrial control panels featuring visual distortions such as blur and masking to test robustness.
On TPanel-UI, Chain-of-Ground outperforms the SOTA MLLM Qwen3-VL-235B by \textbf{6.9\%}, demonstrating the effectiveness of multi-step, training-free grounding across real-world and digital interfaces.
Together, these results point to a new direction for unlocking MLLMs’ grounding potential, through structured, iterative refinement rather than additional training.
\end{abstract}

%% file: sec/1_intro.tex
\begin{figure}[ht]
\centering
  \includegraphics[width=\linewidth]{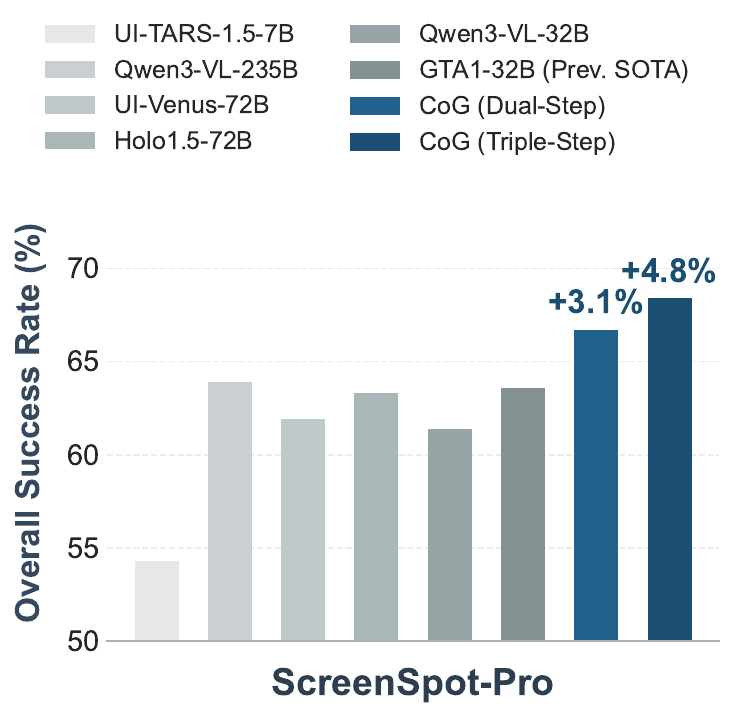}
\caption{
  Overall success rates on the ScreenSpot-Pro~\cite{li2025screenspotpro} benchmark for recent GUI grounding models. 
  GTA1-32B \cite{yang2025gta1guitesttimescaling} represents the previous state of the art among strong baselines including UI-TARS-1.5-7B \cite{qin2025uitarspioneeringautomatedgui}, Qwen3-VL-235B \cite{qwen2025qwen3vl}, UI-Venus-72B \cite{gu2025uivenustechnicalreportbuilding}, Holo1.5-72B \cite{hai2025holo15modelfamily}, and Qwen3-VL-32B. 
  Our Chain-of-Ground (CoG) framework surpasses GTA1-32B by a substantial margin: the dual-step variant yields a +3.1\% improvement, while the triple-step variant achieves a +4.8\% gain, setting a SOTA on ScreenSpot-Pro.
  }
\label{fig:teaser}
\end{figure}

\section{Introduction}

Grounding natural language instructions to UI elements is fundamental for computer-use agents and multimodal systems \cite{seeact,wonderland,kim2024language,mobileagent,seeact,OSWorld,oscopilot,wang2025opencuaopenfoundationscomputeruse,claude-CU,openai2025cua}, yet accuracy remains the main bottleneck to reliability and autonomy. Interfaces pack hundreds of look-alike controls whose meaning depends on structure and function, making precise grounding essential for downstream steps \cite{gou2025uground}. Still, grounding from pixels is hard: targets are small and crowded; icons are compositional; text is tiny or occluded; spatial cues span distant panes; appearance shifts with themes and resolutions; repeats and low contrast mislead; and control boundaries poorly match pixels. Because single-step prediction often misaligns and induces errors, a stepwise procedure that reasons and exposes intermediate decisions yields a more robust and interpretable framework.


\begin{figure*}[!t]
    \centering
    \includegraphics[width=\linewidth]{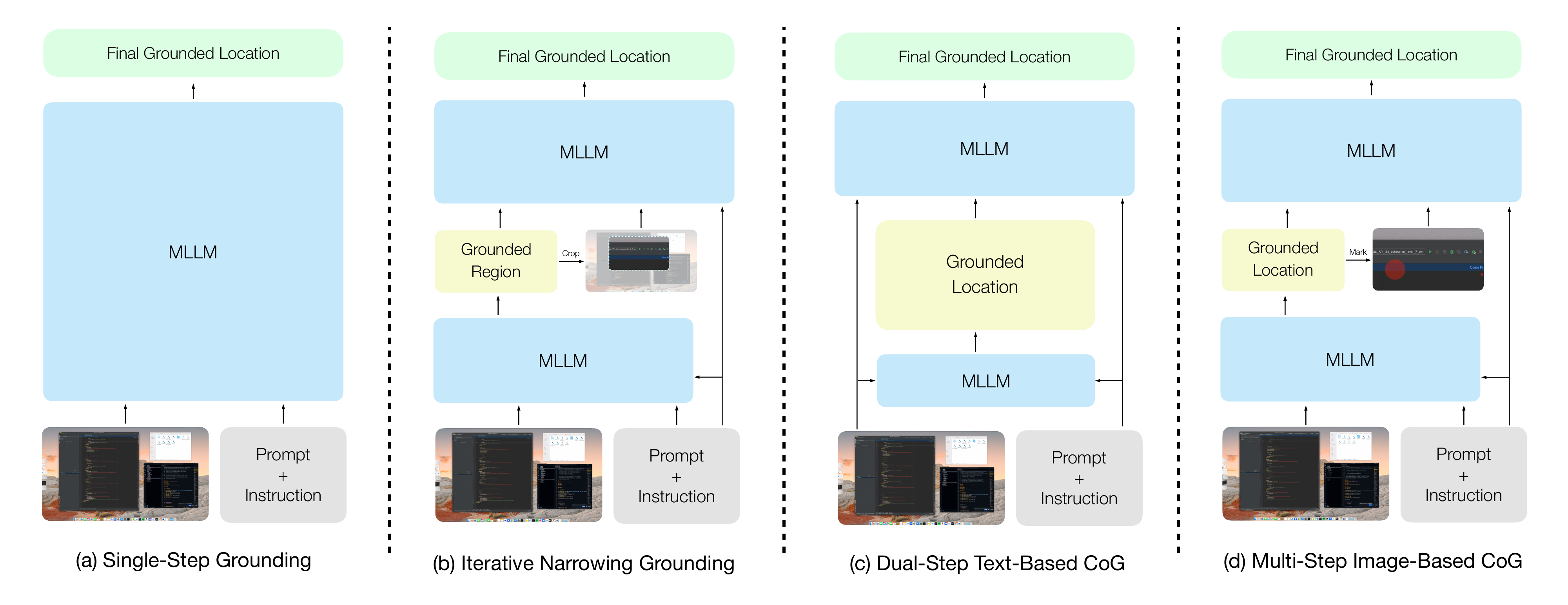}
    \caption{
    Comparison of grounding strategies. We compared previous strategy (a) Single-Step Grounding and (b) Iterative Narrowing Grounding with our proposed Multi-Step Text-Based (c) and Image-Based (d) Chain-of-Ground (CoG).
    (a) Single-Step Grounding performs prediction from screenshot and instruction directly. 
    (b) Iterative Narrowing Grounding crops around the first guess and re-predicts from the cropped image, which can lose global context \cite{nguyen2025improvedguigroundingiterative,wu-etal-2025-dimo}. 
    (c) Multi-Step Text-Based CoG first produces a textual reference and then grounds by reasoning with this reference. 
    (d) Multi-Step Image-Based CoG marks intermediate locations and iteratively refines using visual reference feedback while keeping access to the full screen. 
    }
    \label{fig:overview}
\end{figure*}

Existing GUI grounding methods range from classical detectors and segmentation networks \cite{ren2016fasterrcnnrealtimeobject,ge2021yoloxexceedingyoloseries,liu2023grounding,kirillov2023segment,ronneberger2015unetconvolutionalnetworksbiomedical,lu2024omniparserpurevisionbased} to multimodal LLMs such as GPT-5 \cite{openai2025gpt5}, Qwen3-VL \cite{qwen2025qwen3vl}, Claude Sonnet 4.5 \cite{anthropic2025sonnet45}, Seed 1.5 VL \cite{guo2025seed15vltechnicalreport}, and InternVL3.5 \cite{wang2025internvl35advancingopensourcemultimodal}. While MLLMs are strong at reasoning and grounding, they still struggle with precise localization in cluttered scenes; visual prompting like SoM \cite{som} helps alignment but depends on brittle segmentation, and finetuned MLLMs \cite{cheng2024seeclick,qin2025uitarspioneeringautomatedgui,gou2025uground,yang2025gta1guitesttimescaling,hai2025holo15modelfamily,gu2025uivenustechnicalreportbuilding} improve accuracy yet remain limited in stability and interpretability. Recent iterative localization such as DiMo-GUI \cite{wu-etal-2025-dimo} and Iterative Narrowing \cite{nguyen2025improvedguigroundingiterative} zoom screnshots into cropped regions \emph{(see Fig.~\ref{fig:overview})} but sacrifice global context. 

Motivated by these gaps, we introduce \textbf{Chain-of-Ground (CoG)}, a purely visual framework designed to enhance grounding accuracy through iterative reasoning and reference feedback. CoG first anchors an initial guess, encodes this location as an explicit marker, and then progressively refines the prediction by re-evaluating the instruction in the context of the updated image. The framework produces an interpretable reasoning trace, allowing earlier guesses to be revisited and corrected. It comprises two core components: (1) iterative reasoning, where the model incrementally updates its prediction using both the current hypothesis and prior steps, over a fixed or adaptive number of iterations; and (2) reference feedback, where each predicted location is returned to the model as a visual or textual signal. We design this feedback to balance interpretability and precision, exploring different marker modalities (e.g., image overlays vs. text prompts) and scales (e.g., small vs. large marks).

We evaluate our approach on the ScreenSpot-Pro benchmark, which features high-resolution professional GUIs with complex layouts and dense We evaluate on the ScreenSpot-Pro benchmark with high-resolution professional GUIs. Across multiple MLLM backbones, our reference-guided iterative grounding consistently outperforms direct one-step baselines and sets a new state of the art on the public leaderboard, surpassing the previous best model by 4.1\% in grounding accuracy. The method also reduces prediction instability across domains, and qualitative analysis shows that it reasons relationally by linking targets to contextual references rather than relying on surface-level visual matching.

Although our method achieves strong results on established computer GUI grounding benchmarks, we aim to extend its applicability to real-world industrial interfaces, where conditions are far more challenging. However, existing benchmarks rarely capture such complexity, leaving a gap between academic progress and deployment settings. To bridge this gap, we introduce \textbf{TPanel-UI}, a dataset of 420 labeled instances capturing real industrial control panels such as thermostats, instrument dashboards, and machinery interfaces. Unlike prior GUI datasets focused on virtual or synthetic environments, TPanel-UI emphasizes physical panels with dense layouts, metallic surfaces, variable lighting, and complex iconography representative of real operational settings. Specifically:
(1) it includes 420 high-resolution panel images from 20 commercial brands with diverse layouts;
(2) among them, 100 instances involve physical-button interactions and 320 correspond to touch-based interfaces; and
(3) to stress test robustness, we add controlled degradations at multiple severities including blur, masking, exposure shifts, noise, and compression, yielding paired clean-degraded samples that preserve labels and enable rigorous robustness auditing.
This enables systematic evaluation of grounding robustness and supports research on reference resolution in industry-oriented, safety-critical scenarios.

\vspace{0.1cm}

\noindent
This work makes the following contributions.
\begin{enumerate}
    \item We revisited GUI grounding as a structured reasoning problem rather than a single-step recognition task. To this end, we introduced \textbf{Chain-of-Ground (CoG)},  a reference-guided iterative grounding framework that enables the model to reflect progressively with contextual feedback.  
    \item We proposed three complementary components for the Chain-of-Ground mechanism: a reflective grounding framework that records intermediate hypotheses and allows self-correction, reference guidance that anchors predictions through contextual comparison and semantic analogy, and iterative refinement that progressively reduces ambiguity among visually similar interface elements.  
    \item We presented \textbf{TPanel-UI}, a real industrial control panel dataset for controlled evaluation. 
    \item We verified the effectiveness of our methods on both standard GUI Grounding dataset and real industrial scenarios.

\end{enumerate}


%% file: sec/2_related_works.tex
\section{Related Work}

\subsection{GUI Agents}
Large-scale language and multimodal models \cite{openai2024gpt4ocard,anthropic2025claude,anthropic2025sonnet45,bai2025qwen25vltechnicalreport,openai2025gpt5,google2025gemini2_5} are increasingly used as agents for GUI interaction \citep{wonderland,kim2024language,mobileagent,seeact,OSWorld,oscopilot,wang2025opencuaopenfoundationscomputeruse,claude-CU,openai2025cua}. While many multimodal agents \citep{visualwebarena,webarena,spider} rely on structured data like HTML, our work aligns with research on visually grounded agents operating on raw pixels \citep{shaw2023pixels,autoui,hong2023cogagent,cheng2024seeclick,niu2024screenagent}, proposing a two-stage framework decoupling planning from visual grounding.

\subsection{Visual Grounding}
\paragraph{Visual Detection and Segmentation Methods.} GUI elements can be grounded using classical but brittle techniques or modern detectors \cite{ren2016fasterrcnnrealtimeobject,ge2021yoloxexceedingyoloseries,liu2023grounding} and segmentation networks \cite{kirillov2023segment,ronneberger2015unetconvolutionalnetworksbiomedical}. GUI-specific detectors like OmniParser \cite{lu2024omniparserpurevisionbased} are trained on large GUI corpora.
\vspace{-1em}
\paragraph{General-Purpose Multimodal LLMs.} MLLMs like GPT-5 \cite{openai2025gpt5}, Qwen3-VL \cite{qwen2025qwen3vl}, Clause Sonnet 4.5 \cite{anthropic2025sonnet45}, Seed 1.5 VL \cite{guo2025seed15vltechnicalreport}, and InternVL3.5 \cite{wang2025internvl35advancingopensourcemultimodal} are applied to GUI grounding but can struggle with precise localization. Visual prompting strategies like Set-of-Mark (SoM) \cite{som} are used to improve grounding in systems such as WebVoyager \cite{webvoyager}, VisualWebArena \cite{visualwebarena}, and SeeAct \cite{seeact}, but are limited by segmentation quality.
\vspace{-1em}
\paragraph{Finetuned Multimodal LLMs.} Fine-tuning MLLMs on GUI-specific data improves grounding accuracy. Examples include SeeClick \cite{cheng2024seeclick}, UI-TARS \cite{qin2025uitarspioneeringautomatedgui}, and UGround \cite{gou2025uground}. More recent work uses reinforcement learning, such as GTA1-32B \cite{yang2025gta1guitesttimescaling}, Holo 1.5 \cite{hai2025holo15modelfamily}, and UI-Venus \cite{gu2025uivenustechnicalreportbuilding}. Despite advances, even state-of-the-art models like GTA1-32B achieve only 63.6\% accuracy on ScreenSpot-Pro \cite{li2025screenspotpro}, showing persistent challenges.
\vspace{-1em}
\paragraph{Iterative Grounding}
Recent work has explored iterative localization. Zoom-based methods like DiMo-GUI~\citep{wu-etal-2025-dimo}, Iterative Narrowing~\citep{nguyen2025improvedguigroundingiterative}, and GUI-Spotlight~\citep{lei2025textscguispotlightadaptiveiterativefocus} refine predictions by cropping image regions. However, these methods lose global context. Our framework iterates in the reasoning space, preserving global context and allowing the model to revisit and refine its hypotheses through an evolving reasoning trace with referential feedback.

\begin{figure*}[t!]
    \centering
    \includegraphics[width=\linewidth]{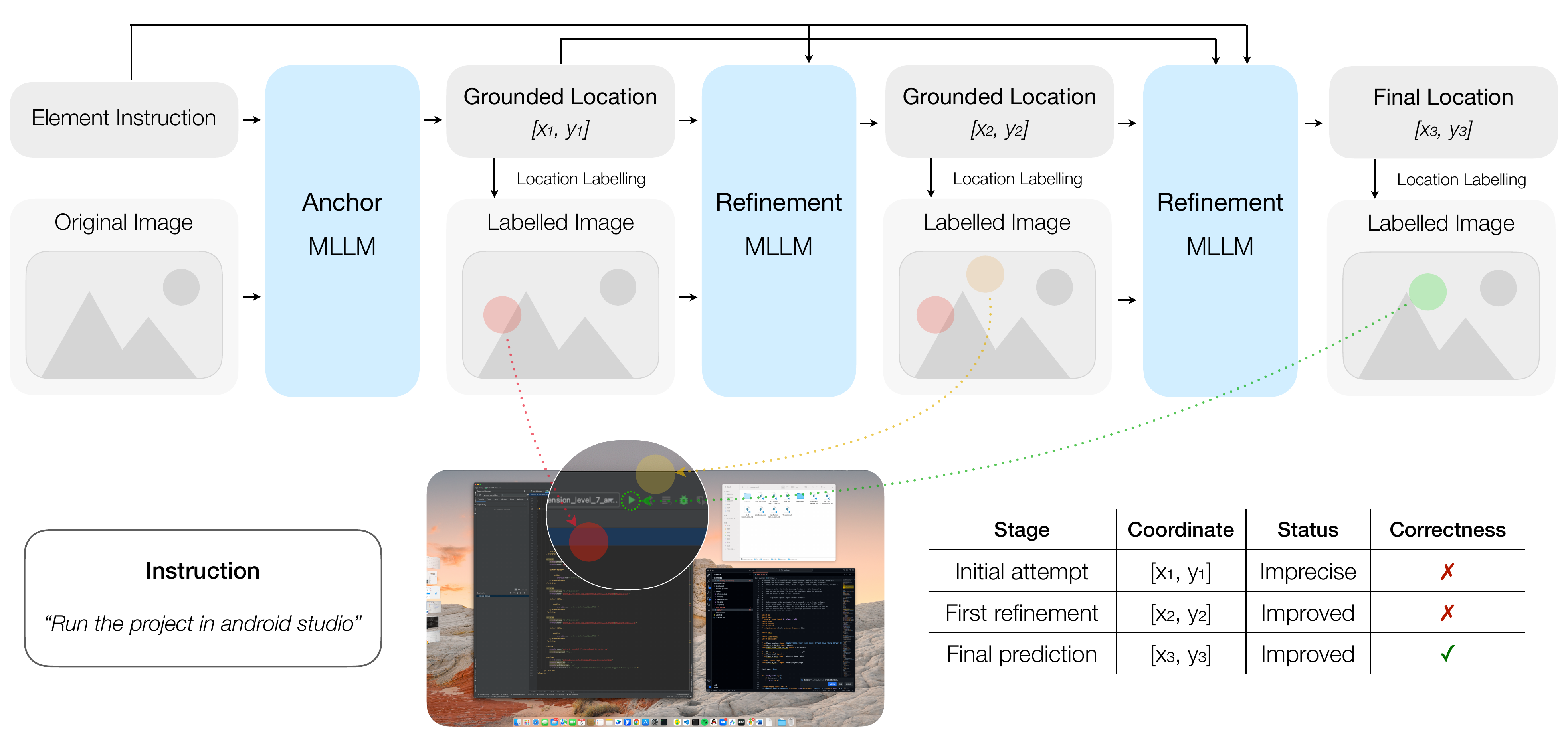}
    \caption{
    Overview of the \textbf{Chain-of-Ground (CoG)} framework, shown here in its triple-step configuration. 
    Given an element instruction and the original interface image, CoG performs grounding through a sequence of iterative reasoning steps using multimodal large language models (MLLMs). 
    The first Anchor MLLM predicts an initial location, which is labeled on the image and passed forward as contextual feedback. 
    Two subsequent Refinement stages reflect the instruction in relation to the updated visual context and iteratively refine the coordinates. 
    This structured reasoning loop terminates at the final grounded location $[x_3, y_3]$. 
    By leveraging multi-step visual feedback without model retraining, CoG enables accurate, interpretable, and training-free grounding in complex GUI environments.
    }
    \label{fig:architecture}
\end{figure*}

%% file: sec/3_methods.tex
\section{Methods}

Our proposed \textbf{Chain-of-Ground (CoG)} is a GUI grounding framework that performs multi-step refinement instead of a single forward prediction. At each step the model revisits the screenshot with cues derived from its previous output, enabling verification and correction. CoG consists of:
(1) \textit{Iterative reasoning}: a fixed or adaptive number of steps where the model re-encodes the screen and updates the target location based on the current hypothesis and referenced history.
(2) \textit{Reference feedback}: the previous prediction is fed back as an input signal to guide the next step. We study design choices for this signal, including modality (text vs. image label) and visual marker size (small vs. large).

\subsection{Iterative Reasoning}
In CoG, grounding proceeds through a sequence of steps rather than a single forward pass. The process begins with an anchoring step, where the model produces an initial localization based on the instruction and the raw screenshot. This anchor serves as an initial hypothesis of the target region. Subsequent steps then refine this anchor through iterative updates: each step encodes previous outputs into a feedback signal and uses it, together with the original instruction, to generate an improved prediction. Repeating this reflect–refine loop allows the model to self-correct and gradually converge toward a stable and accurate grounding.

We evaluate three configurations to study the role of iterative depth: a single-step baseline, a dual-step setup with one refinement round, and a triple-step setup with two refinement rounds. This allows us to measure the benefits of reflective correction. Furthermore, recognizing that different models may excel at different stages of reflective reasoning, we also explore various model combinations within the multi-step framework to see if leveraging diverse model strengths can further enhance performance.

Across all settings, each iteration involves three conceptual phases: (1) anchoring, which produces an initial hypothesis; (2) reference construction, which encodes prior predictions as feedback; and (3) refinement, which integrates this feedback to reflectively update the localization. This iterative design yields a transparent reasoning trace that captures how the model’s spatial understanding evolves over time.

\subsection{Reference Feedback}
The reference feedback mechanism determines how the model perceives and reuses its previous predictions. Its goal is to provide an explicit signal that helps the model align new reasoning with past observations.  We consider two feedback modalities. In the text-based method, the model encodes its earlier predicted location as a textual coordinate appended to the instruction. This form provides an abstracted signal that can be interpreted by models for a rough reference. In the image-based method, the model receives the marked screenshot with a visual marker on previously predicted locations. The feedback acts as an anchor that allows the model to understand the spatial coordinate of the previous prediction across iterations and also the contextual dependence between UI elements. 
\vspace{1em}
\begin{figure}[h]
    \centering
    \includegraphics[width=\linewidth]{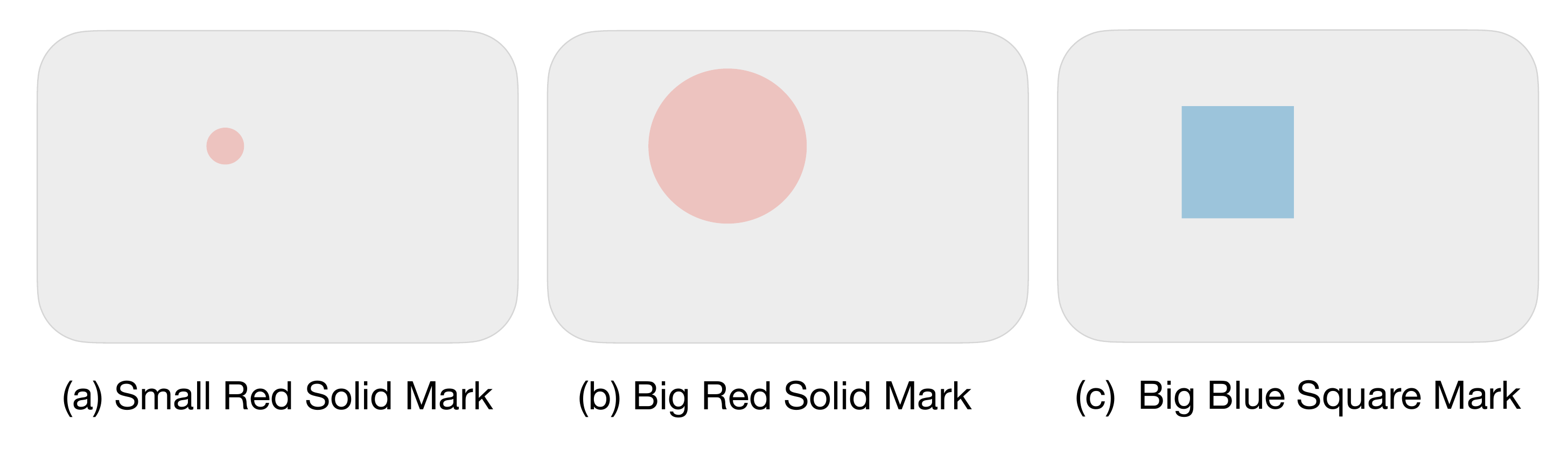}
    \caption{
    Examples of visual grounding marks. (a) A small red solid mark with radius $r=10$. (b) A large red solid mark with radius $r=100$. (c) A large blue square mark with side $l=100$.
    }
    \label{fig:markers}
\end{figure}
For image-based feedback, as shown in Figure~\ref{fig:markers}, we examine how the marker design influences grounding. The main factor is marker size. On high-resolution screenshots, a larger marker is more salient and reliably draws attention to the focus region, whereas very small markers may be overlooked by the model. However, excessively large markers can still occlude nearby elements. In our triple-step mechanism, we used a large red circle \textit{(r=100px)} for the first layer and a large blue square \textit{(l=100px)} for the second step.
By integrating feedback into the reasoning process, CoG encourages the model to revisit earlier hypotheses with additional context. The result is a structured form of grounding where predictions are not isolated outcomes but part of a coherent sequence guided by reflective, interpretable feedback.

%% file: sec/4_tpanel-ui.tex
\section{TPanel-UI Dataset}

Existing GUI grounding datasets such as ScreenSpot-Pro \cite{li2025screenspotpro} and Point Arena \cite{cheng2025pointarenaprobingmultimodalgrounding} mainly cover virtual interfaces or generic photography, leaving real hardware panels underexplored. \textbf{TPanel-UI} addresses this gap with industrial-style scenes such as thermostats and instrument dashboards, where controls are small, densely packed, and often icon-based. The dataset contains \textit{420} high-resolution panel images spanning \textit{20} commercial brands with diverse layouts. Among these, \textit{100} instances involve physical-button interactions, while \textit{320} correspond to touch-based interfaces. For each image, annotators identify a single interactive control element (e.g., power button) required to fulfill a given instruction and record its precise coordinate. Each image is paired with a concise natural-language instruction describing the intended user action, supporting controlled evaluation of grounding performance under sparse linguistic guidance and high visual ambiguity.

\begin{figure}[h]
    \centering
    \includegraphics[width=\linewidth]{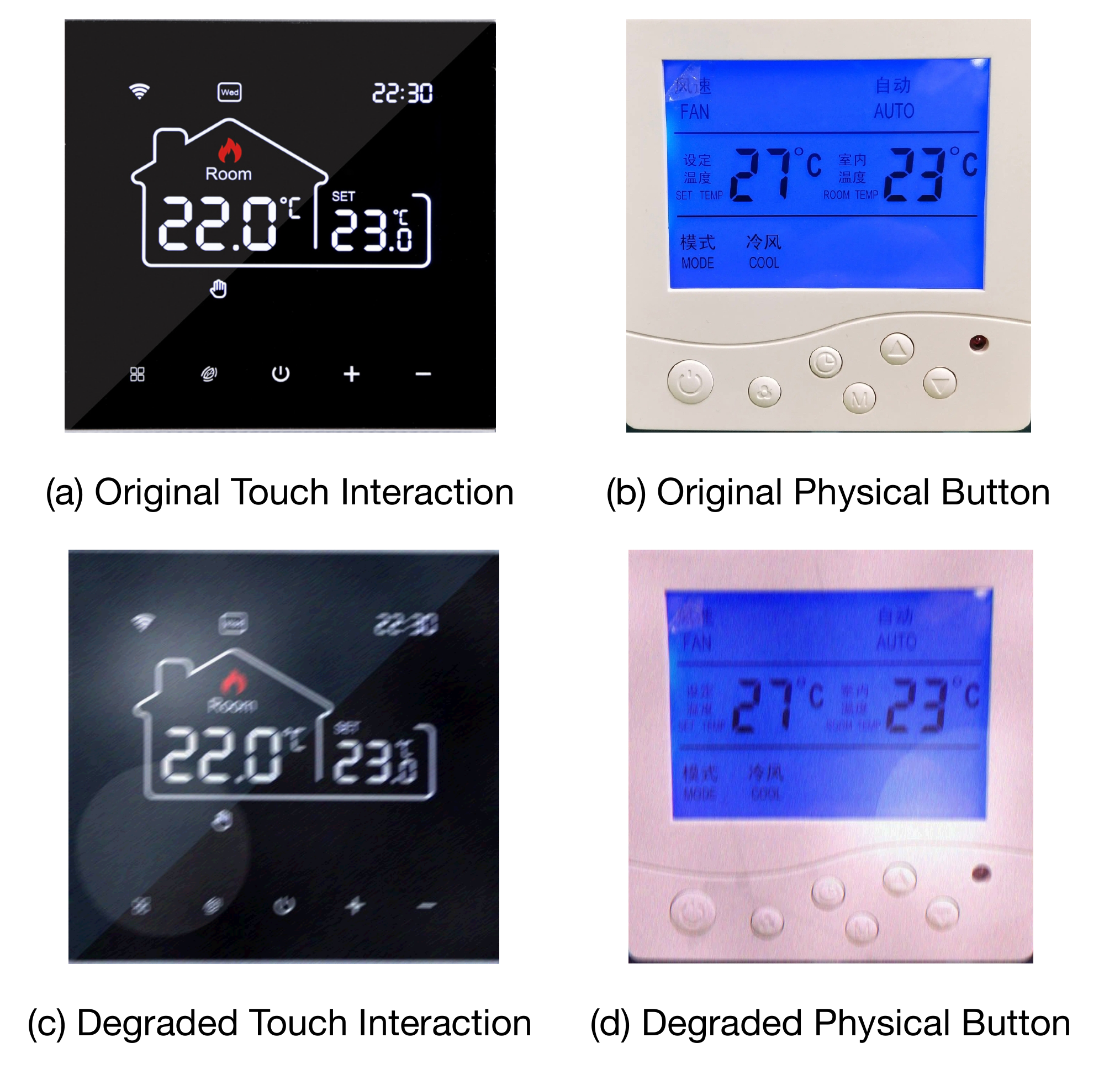}
    \caption{
    Examples from the TPanel-UI dataset. 
    The dataset includes two interface types: (a,c) touch-based control panels and (b,d) physical-button panels. 
    For each, we provide both original images and degraded versions featuring blur, glare, and lighting artifacts to assess grounding robustness under visually challenging conditions.
    }
    \label{fig:tpanel-ui}
\end{figure}

Images are captured using consumer-grade cameras under everyday usage conditions, then processed through a multi-stage degradation pipeline to simulate realistic visual disturbances. Starting from the original photos, we apply randomized transformations including global color shifts, synthetic lens flares, Perlin-noise overlays, and wear patterns to emulate surface damage. Additional degradations such as Gaussian sensor noise, motion blur, soft specular highlights, and chromatic aberration introduce further distortion. JPEG recompression and color quantization reduce image fidelity, while lighting inconsistencies, mild occlusions, and perspective shifts add complexity. The result is a challenging set of degraded variants corresponding to the original panel images. 

In summary, TPanel-UI provides 420 annotated panel instances (original and degraded) sharing the same interaction targets. This design yields a controlled benchmark well-suited for evaluating model robustness and the reliability of iterative grounding under realistic visual noise for industrial scene.

%% file: sec/5_eval.tex
\section{Experiments}
\renewcommand{\arraystretch}{1.15}
\begin{table*}[!t]
\setlength{\tabcolsep}{4pt}
\centering
\caption{
Comparison with state-of-the-art methods on the ScreenSpot-Pro dataset \cite{li2025screenspotpro}. We report grounding accuracy (\%) by domain and show only the averaged scores (Avg). A dash `--' indicates unavailable results and `$^*$' denotes our own evaluation. Our multi-step CoG framework allows flexible model selection per step. The dual-step configuration shown here uses two instances of \textit{Qwen3-VL-235B} \cite{qwen2025qwen3vl}, while the triple-step configuration (illustrated) adopts one possible sequence: \textit{UI-TARS1.5-7B} $\rightarrow$ \textit{Qwen3-VL-235B} $\rightarrow$ \textit{Qwen3-VL-32B}.
}
\small
\vspace{-0.75em}
\resizebox{0.8\textwidth}{!}{%
\begin{tabularx}{\textwidth}{
@{}l
*{7}{>{\centering\arraybackslash}X}
@{}}
\toprule
\textbf{Agent Model} & \textbf{Development} & \textbf{Creative} & \textbf{CAD} & \textbf{Scientific} & \textbf{Office} & \textbf{OS} & \textbf{Avg} \\
\midrule
\rowcolor{gray!15}
\multicolumn{8}{l}{\textit{Proprietary Models}} \\
GPT-4o \cite{openai2024gpt4ocard} & 0.7 & 0.6 & 1.5 & 1.2 & 0.9 & 0.0 & 0.8 \\
Claude 3.7 Sonnet \cite{anthropic2025claude} & - & - & - & - & - & - & 27.7 \\
Operator \cite{openai2025cua} & 35.1 & 39.6 & 16.1 & 43.7 & 53.0 & 32.7 & 36.6 \\
Seed-1.5-VL \cite{guo2025seed15vltechnicalreport} & 53.8 & 59.2 & 59.0 & 61.4 & 74.8 & 60.2 & 60.9 \\
UI-TARS-1.5 \cite{qin2025uitarspioneeringautomatedgui} & 63.9 & 50.4 & 58.2 & \underline{69.3} & 79.6 & 51.0 & 61.6 \\

\rowcolor{gray!15}
\multicolumn{8}{l}{\textit{Open-Source Models}} \\
SeeClick \cite{cheng2024seeclick} & 0.3 & 0.6 & 1.9 & 2.0 & 0.9 & 1.5 & 1.1 \\
CogAgent-18B \cite{Hong_2024_CVPR} & 8.0 & 5.6 & 6.1 & 13.4 & 10.0 & 3.1 & 7.7 \\
Aria-UI \cite{yang2025ariauivisualgroundinggui} & 8.4 & 14.7 & 6.1 & 18.1 & 16.1 & 2.6 & 11.3 \\
OS-Atlas-7B \cite{wu2024atlas} & 17.7 & 17.9 & 10.3 & 24.4 & 27.4 & 16.8 & 18.9 \\
Qwen2.5-VL-7B \cite{bai2025qwen25vltechnicalreport} & 29.1 & 24.9 & 13.8 & 31.1 & 45.7 & 22.4 & 27.6 \\
UGround-72B \cite{gou2025uground} & 31.1 & 35.8 & 13.8 & 50.0 & 51.3 & 25.5 & 34.5 \\
SE-GUI-3B \cite{yuan2025enhancingvisualgroundinggui} & 35.1 & 29.0 & 31.8 & 43.3 & 50.9 & 25.5 & 35.9 \\
Jedi-3B \cite{xie2025scalingcomputerusegroundinguser} & 38.1 & 34.6 & 23.0 & 38.6 & 57.0 & 25.0 & 36.1 \\
GUI-G1-3B \cite{zhou2025guig} & 31.1 & 26.6 & 32.2 & 48.0 & 59.1 & 16.1 & 37.1 \\
UI-TARS-72B \cite{qin2025uitarspioneeringautomatedgui} & 40.8 & 39.6 & 17.2 & 45.7 & 54.8 & 30.1 & 38.1 \\
Jedi-7B \cite{xie2025scalingcomputerusegroundinguser} & 27.4 & 34.0 & 32.2 & 52.4 & 68.7 & 26.0 & 39.5 \\
UI-TARS-1.5-7B \cite{qin2025uitarspioneeringautomatedgui} & 31.8 & 40.2 & 31.8 & 47.2 & 65.6 & 33.2 & 42.0 \\
SE-GUI-7B \cite{yuan2025enhancingvisualgroundinggui} & 44.5 & 37.2 & 42.1 & 54.7 & 70.4 & 38.8 & 47.3 \\
GUI-Spotlight \cite{lei2025textscguispotlightadaptiveiterativefocus} & 52.6 & 44.8 & 51.6 & 53.5 & 70.4 & 45.9 & 52.8 \\
Qwen2.5-VL-72B-Instruct  \cite{bai2025qwen25vltechnicalreport} & 53.5 & 44.9 & 44.4 & 59.1 & 72.6 & 49.5 & 53.3 \\
Qwen3-VL-32B-Instruct$^*$ \cite{qwen2025qwen3vl}& \underline{67.5} & 57.6 & 47.9 & 65.3 & 73.6 & 60.1 & 61.4 \\
Holo1.5-72B \cite{hai2025holo15modelfamily} & 63.5 & 62.5 & 51.3 & 64.2 & 79.6 & 59.7 & 63.3 \\
GTA-32B \cite{yang2025gta1guitesttimescaling} & 61.2 & 52.8 & \textbf{60.5} & 65.0 & \textbf{83.5} & \underline{65.3} & 63.6 \\
Qwen3-VL-235B-Instruct$^*$ \cite{qwen2025qwen3vl}& 66.4 & \underline{63.1} & 52.8 & 66.1 & 75.8 & 61.9 & 63.9 \\

\rowcolor{gray!15}
\multicolumn{8}{l}{\textit{Ours (Multi-Step CoG)}} \\
\textbf{Dual-Step CoG}$^*$ & 66.4 & \underline{63.1} & \underline{59.5} & \textbf{70.9} & \underline{80.9} & 62.3 & \underline{66.7} \\
\textbf{Triple-Step CoG}$^*$ & \textbf{69.2} & \textbf{65.7} & 57.9 & \underline{69.3} & \underline{80.9} & \textbf{71.9} & \textbf{68.4} \\
\bottomrule
\end{tabularx}
}

\label{tab:screenspot_pro_avg_only}
\end{table*}
We evaluate our Chain-of-Ground (CoG) framework by examining how reasoning depth, feedback modality, and visual design affect grounding accuracy. Our study uses large-scale GUI datasets and a new industrial benchmark to test performance under varied conditions. We evaluate on two primary benchmarks:  
(1) \textbf{ScreenSpot-Pro} \cite{li2025screenspotpro}, a large-scale dataset for grounding generalization in professional GUI environments (see Table~\ref{tab:screenspot_pro_avg_only}).
(2) \textbf{TPanel-UI}, a new 420-label dataset of industrial control panels with visual distortions to test real-world robustness (see Table~\ref{tab:tpanel_ui_results}).

\subsection{ScreenSpot-Pro Results}
For ScreenSpot-Pro datasets , test scenes cover six major categories of professional software: \textit{Development}, \textit{Creative}, \textit{CAD}, \textit{Scientific}, \textit{Office}, and \textit{Operating Systems}, representing a diverse range of interface structures and visual densities (see Table~\ref{tab:screenspot_pro_avg_only}). We test CoG on several MLLMs of different scales and capabilities, including Qwen-VL-32B, Qwen-VL-235B \cite{qwen2025qwen3vl}, and UI-TARS-1.5-7B \cite{qin2025uitarspioneeringautomatedgui}. Performance is measured using grounding success rate across categories. Our results demonstrate that our multi-step CoG framework achieves state-of-the-art performance, outperforming existing proprietary and open-source models.
Specifically, our triple-step CoG model achieves an average accuracy of \textit{68.4\%}, surpassing the strongest open-source model, GTA-32B (63.6\%), by a significant margin of \textit{4.8\%}.
Our dual-step CoG model also shows strong performance with an average accuracy of \textit{66.7\%}, outperforming GTA-32B by \textit{3.1\%}.
Notably, our triple-step model is the top performer across most categories, including \textit{Development} (69.2\%), \textit{Creative} (65.7\%), and \textit{OS} (71.9\%).
The dual-step CoG model achieves the best results in the \textit{Scientific} (70.9\%) and \textit{Office} (80.9\%) categories, while the single-step Qwen3-VL-235B model is the top performer in the \textit{CAD} (59.5\%) category.
These findings highlight the effectiveness of our iterative refinement approach in improving grounding accuracy, especially in challenging professional software environments.

\subsection{TPanel-UI Results}

On the \textbf{TPanel-UI} dataset, which features challenging industrial control panels, our CoG framework again demonstrates significant performance gains. For these experiments, visual feedback was provided using a small red circle to mark the predicted location in each step. As shown in Table~\ref{tab:tpanel_ui_results}, all dual-step CoG configurations outperform their single-step counterparts. The heterogeneous combination of Qwen3-VL-235B $\rightarrow$ Qwen3-VL-32B achieves the highest average accuracy of 90.0\%, a 6.9\% improvement over the strongest single model (Qwen3-VL-235B). Furthermore, with the same refinement model as anchoring model, with Qwen3-VL-32B $\rightarrow$ Qwen3-VL-32B also yields a substantial boost, improving accuracy from 81.2\% to 87.9\%. These results underscore CoG's robustness and its effectiveness in leveraging iterative refinement to overcome visual complexities in specialized domains.

\begin{table*}[h!]
\setlength{\tabcolsep}{4pt}
\renewcommand{\arraystretch}{1.15}
\centering
\caption{Performance on the TPanel-UI dataset. We report single-step baselines and dual-step CoG runs. Dual-step settings include Qwen3-VL-32B followed by Qwen3-VL-32B, and vice versa. The Qwen3-VL-235B $\rightarrow$ Qwen3-VL-32B setting achieves the best results.}
\small
\resizebox{0.8\textwidth}{!}{%
\begin{tabularx}{\textwidth}{
@{}l
*{3}{>{\centering\arraybackslash}X}
@{}}
\toprule
\textbf{Method} & \textbf{Touch Interaction} & \textbf{Physical Button} & \textbf{Avg} \\
\midrule
\rowcolor{gray!15}
\multicolumn{4}{l}{\textit{Single Models}} \\
Qwen3-VL-32B & 82.2 & 74.0 & 81.2 \\
Qwen3-VL-235B & \underline{84.7} & \underline{78.0} & 83.1 \\
UI-TARS-1.5-7B & 50.0 & 40.0 & 47.6 \\
\midrule
\rowcolor{gray!15}
\multicolumn{4}{l}{\textit{Dual-Step CoG}} \\
CoG (Qwen3-VL-235B $\rightarrow$ Qwen3-VL-235B) & 83.6 & \underline{78.0} & 84.3 \\
CoG (Qwen3-VL-32B $\rightarrow$ Qwen3-VL-32B) & \textbf{90.9} & \underline{78.0} & \underline{87.9} \\
CoG (Qwen3-VL-235B $\rightarrow$ Qwen3-VL-32B) & \textbf{90.9} & \textbf{87.0} & \textbf{90.0} \\
CoG (UI-TARS-1.5-7B $\rightarrow$ UI-TARS-1.5-7B) & 64.4 & 54.0 & 61.9 \\
\bottomrule
\end{tabularx}
}
\label{tab:tpanel_ui_results}
\end{table*}

\subsection{Ablation Studies}

\paragraph{Number of Iterations. }
Table~\ref{tab:ablation_iterations} shows the impact of the number of reasoning iterations on performance. We observe a consistent improvement in accuracy as the number of iterations increases from one to three. The single-step model achieves a baseline accuracy of 63.9\%. The dual-step model improves this by 2.8\% to 66.7\%. The triple-step model further boosts accuracy to 68.4\%, a 4.5\% improvement over the single-step baseline. This trend highlights the value of iterative refinement in our CoG framework, allowing the model to progressively correct errors and improve grounding accuracy.

\begin{table}[t]
\setlength{\tabcolsep}{4pt}
\renewcommand{\arraystretch}{1.15}
\centering
\caption{Ablation on the number of reasoning iterations on the ScreenSpot-Pro dataset \cite{li2025screenspotpro}. 2-step CoG uses Qwen3-VL-235B in both steps, while 3-step CoG uses the sequence UI-TARS1.5-7B $\rightarrow$ Qwen3-VL-235B $\rightarrow$ Qwen3-VL-32B.}
\small
\vspace{-0.75em}
\begin{tabularx}{\columnwidth}{
@{}l
*{7}{>{\centering\arraybackslash}X}
@{}}\toprule
\textbf{Method} & \textbf{Dev} & \textbf{Crea} & \textbf{CAD} & \textbf{Sci} & \textbf{Office} & \textbf{OS} & \textbf{Avg} \\
\midrule
\rowcolor{gray!15}
\multicolumn{8}{l}{\textit{Baselines (Single-Step)}} \\
UI-TARS-1.5-7B & 31.8 & 40.2 & 31.8 & 47.2 & 65.6 & 33.2 & 42.0 \\
Qwen3-VL-32B & \underline{67.5} & 57.6 & 47.9 & 65.3 & 73.6 & 60.1 & 61.4 \\
Qwen3-VL-235B & 66.4 & \underline{63.1} & 52.8 & \underline{66.1} & \underline{75.8} & 61.9 & 63.9 \\
\midrule
\rowcolor{gray!15}
\multicolumn{8}{l}{\textit{CoG Iterations}} \\
Dual-Step CoG & 66.4 & \underline{63.1} & \textbf{59.5} & \textbf{70.9} & \textbf{80.9} & \underline{62.3} & \underline{66.7} \\
Triple-Step CoG & \textbf{69.2} & \textbf{65.7} & \underline{57.9} & \textbf{70.9} & \textbf{80.9} & \textbf{71.9} & \textbf{68.4} \\
\bottomrule
\end{tabularx}
\label{tab:ablation_iterations}
\end{table}
\vspace{-1em}
\paragraph{Feedback Modality.}
We investigate how different feedback modalities affect performance in the CoG framework. As shown in Table~\ref{tab:ablation_feedback_modality}, visual-only feedback achieves the highest accuracy of 65.8\%, while text-only feedback results in a lower accuracy of 64.3\%. Removing feedback entirely yields the worst performance at 61.4\%. These findings indicate that visual markers provide more effective grounding cues than textual descriptions alone, though both contribute positively relative to no feedback. Effective grounding in our multi-step setup benefits from spatially explicit guidance.

\begin{table}[t]
\setlength{\tabcolsep}{4pt}
\renewcommand{\arraystretch}{1.15}
\centering
\caption{Ablation on feedback modality on the ScreenSpot-Pro dataset \cite{li2025screenspotpro}. Both text- and image-based feedback designs use the Qwen3-VL-32B model in both steps.}
\small
\vspace{-0.75em}
\begin{tabularx}{\columnwidth}{
@{}l
*{7}{>{\centering\arraybackslash}X}
@{}}\toprule
\textbf{Feedback Design} & \textbf{Dev} & \textbf{Crea} & \textbf{CAD} & \textbf{Sci} & \textbf{Office} & \textbf{OS} & \textbf{Avg} \\
\midrule
\rowcolor{gray!15}
\multicolumn{8}{l}{\textit{Baselines (Single-Step)}} \\
Qwen3-VL-32B & \textbf{67.5} & 57.6 & 47.9 & 65.3 & 73.6 & 60.1 & 61.4 \\
\midrule
\rowcolor{gray!15}
\multicolumn{8}{l}{\textit{Modality}} \\
Text-based & 65.1 & \underline{62.7} & \underline{54.6} & \underline{67.3} & \underline{77.0} & \textbf{61.7} & \underline{64.3} \\
Image-based & \underline{65.4} & \textbf{64.7} & \textbf{57.2} & \textbf{69.3} & \textbf{79.6} & \underline{61.2} & \textbf{65.8} \\
\bottomrule
\end{tabularx}
\label{tab:ablation_feedback_modality}
\end{table}
\vspace{-1em}
\paragraph{Visual Marker Size. }
In Table~\ref{tab:ablation_marker_size}, we analyze the effect of visual marker size on grounding accuracy. The results show that using a large marker (r=100px) leads to a slightly better accuracy of 66.7\% compared to a small marker (r=10px), which achieves 65.5\%. While both marker sizes perform well, the large marker's slight advantage suggests that a more prominent visual cue can help the model better localize the target. However, the small performance difference indicates that a less conspicuous marker is still highly effective, demonstrating flexibility in marker size selection.

\begin{table}[h!]
\setlength{\tabcolsep}{4pt}
\renewcommand{\arraystretch}{1.15}
\centering
\caption{Ablation on visual marker size for image-based feedback on the ScreenSpot-Pro dataset \cite{li2025screenspotpro}. Both marker sizes are tested with the \textbf{Qwen3-VL-32B model} in a dual-step CoG setup.}
\small
\vspace{-0.75em}
\begin{tabularx}{\columnwidth}{
@{}l
*{7}{>{\centering\arraybackslash}X}
@{}}\toprule
\textbf{Marker Size} & \textbf{Dev} & \textbf{Crea} & \textbf{CAD} & \textbf{Sci} & \textbf{Office} & \textbf{OS} & \textbf{Avg} \\
\midrule
\rowcolor{gray!15}
\multicolumn{8}{l}{\textit{Baselines (Single-Step)}} \\
Qwen3-VL-32B & \textbf{67.5} & 57.6 & 47.9 & 65.3 & 73.6 & 60.1 & 61.4 \\
\midrule
\rowcolor{gray!15}
\multicolumn{8}{l}{\textit{Marker Size (Qwen3-VL-32B $\rightarrow$ Qwen3-VL-32B)}} \\
Small Mark & 64.7 & \underline{63.7} & 55.9 & 67.3 & 78.9 & \textbf{62.3} & 65.0 \\
Large Mark & 65.4 & \textbf{64.7} & 57.2 & 69.3 & \underline{79.6} & \underline{61.2} & \underline{65.8} \\
\midrule
\rowcolor{gray!15}
\multicolumn{8}{l}{\textit{Marker Size (Qwen3-VL-235B $\rightarrow$ Qwen3-VL-32B)}} \\
Small Mark & \underline{66.7} & 61.4 & \underline{57.9} & \textbf{72.0} & 76.9 & 60.2 & 65.5 \\
Large Mark & 66.4 & 63.1 & \textbf{59.5} & \underline{70.9} & \textbf{80.9} & \textbf{62.3} & \textbf{66.7} \\
\bottomrule
\end{tabularx}
\label{tab:ablation_marker_size}
\end{table}

\begin{table*}[h!]
\setlength{\tabcolsep}{4pt}
\renewcommand{\arraystretch}{1.15}
\centering
\caption{Ablation on model combination for triple-step CoG on the ScreenSpot-Pro dataset \cite{li2025screenspotpro}.}
\small
\vspace{-0.75em}
\resizebox{0.8\textwidth}{!}{%
\begin{tabularx}{\textwidth}{
@{}l
*{7}{>{\centering\arraybackslash}X}
@{}}\toprule
\textbf{Model Combination} & \textbf{Dev} & \textbf{Creative} & \textbf{CAD} & \textbf{Scientific} & \textbf{Office} & \textbf{OS} & \textbf{Avg} \\
\midrule
\rowcolor{gray!15}
\multicolumn{8}{l}{\textit{Baselines (Single-Step)}} \\
UI-TARS-1.5-7B & 31.8 & 40.2 & 31.8 & 47.2 & 65.6 & 33.2 & 42.0 \\
Qwen3-VL-32B & 67.5 & 57.6 & 47.9 & 65.3 & 73.6 & 60.1 & 61.4 \\
Qwen3-VL-235B & 66.4 & 63.1 & 52.8 & 66.1 & 75.8 & 61.9 & 63.9 \\
\midrule
\rowcolor{gray!15}
\multicolumn{8}{l}{\textit{Dual-Step CoG Variants}} \\
Qwen3-VL-32B $\rightarrow$ Qwen3-VL-32B & 65.4 & 64.7 & 57.2 & 69.3 & \underline{79.6} & 61.2 & 65.8 \\
Qwen3-VL-235B $\rightarrow$ Qwen3-VL-32B & 66.4 & 63.1 & \underline{59.5} & \textbf{70.9} & \textbf{80.9} & \underline{62.3} & 66.7 \\
\midrule
\rowcolor{gray!15}
\multicolumn{8}{l}{\textit{Triple-Step CoG Variants}} \\
Qwen3-VL-235B $\rightarrow$ Qwen3-VL-32B $\rightarrow$ Qwen3-VL-235B & 69.2 & 60.8 & \textbf{61.8} & \textbf{70.9} & 76.6 & 60.7 & 66.4 \\
Qwen3-VL-32B $\rightarrow$ Qwen3-VL-32B $\rightarrow$ Qwen3-VL-32B & 69.8 & \textbf{66.3} & 58.5 & 68.5 & 78.3 & 66.3 & \underline{67.5} \\
UI-TARS-1.5-7B $\rightarrow$ Qwen3-VL-235B $\rightarrow$ Qwen3-VL-32B & 69.2 & \underline{65.7} & 57.9 & 69.3 & \textbf{80.9} & \textbf{71.9} & \textbf{68.4} \\
Qwen3-VL-235B $\rightarrow$ Qwen3-VL-235B $\rightarrow$ Qwen3-VL-235B & \underline{69.9} & 62.4 & 56.9 & \underline{69.7} & 77.0 & 59.2 & 65.6 \\
Qwen3-VL-32B $\rightarrow$ UI-TARS-1.5-7B $\rightarrow$ Qwen3-VL-32B & 63.4 & 62.7 & 59.2 & 67.3 & 74.3 & 60.2 & 64.3 \\
Qwen3-VL-32B $\rightarrow$ Qwen3-VL-235B $\rightarrow$ Qwen3-VL-32B & \textbf{70.9} & 64.7 & 52.6 & \textbf{70.9} & 78.3 & \underline{66.9} & 66.7 \\
\bottomrule
\end{tabularx}
}
\label{tab:ablation_model_combination}
\end{table*}

\paragraph{Model Combination. }
We investigate the impact of different model combinations in the multi-step CoG framework. Different models may have unique strengths and weaknesses, or “blindspots,” when interpreting visual information. By combining models, we can potentially compensate for these individual limitations and achieve a more robust overall performance. As shown in Table~\ref{tab:ablation_model_combination}, the UI-TARS-1.5-7B $\rightarrow$ Qwen3-VL-235B $\rightarrow$ Qwen3-VL-32B combination achieves the best performance, with an average accuracy of 68.4\%. Based on this, we speculate that the best results are achieved by combining three different models. Although UI-TARS-1.5-7B has the lowest score as a standalone model, we observe that greater model diversity in the combination leads to better performance. We hypothesize that this is because each model has its own deficiencies, and the different decision-making processes of the models compensate for these shortcomings. This suggests that leveraging a smaller, specialized model for the initial step, followed by larger, more capable models for refinement, is an effective strategy. The performance of other combinations, while slightly lower, is still competitive, with the Qwen3-VL-32B $\rightarrow$ Qwen3-VL-235B $\rightarrow$ Qwen3-VL-32B combination, for example, achieving a 5.3\% improvement over the baseline.

\begin{figure}
    \centering
    \includegraphics[width=1\linewidth]{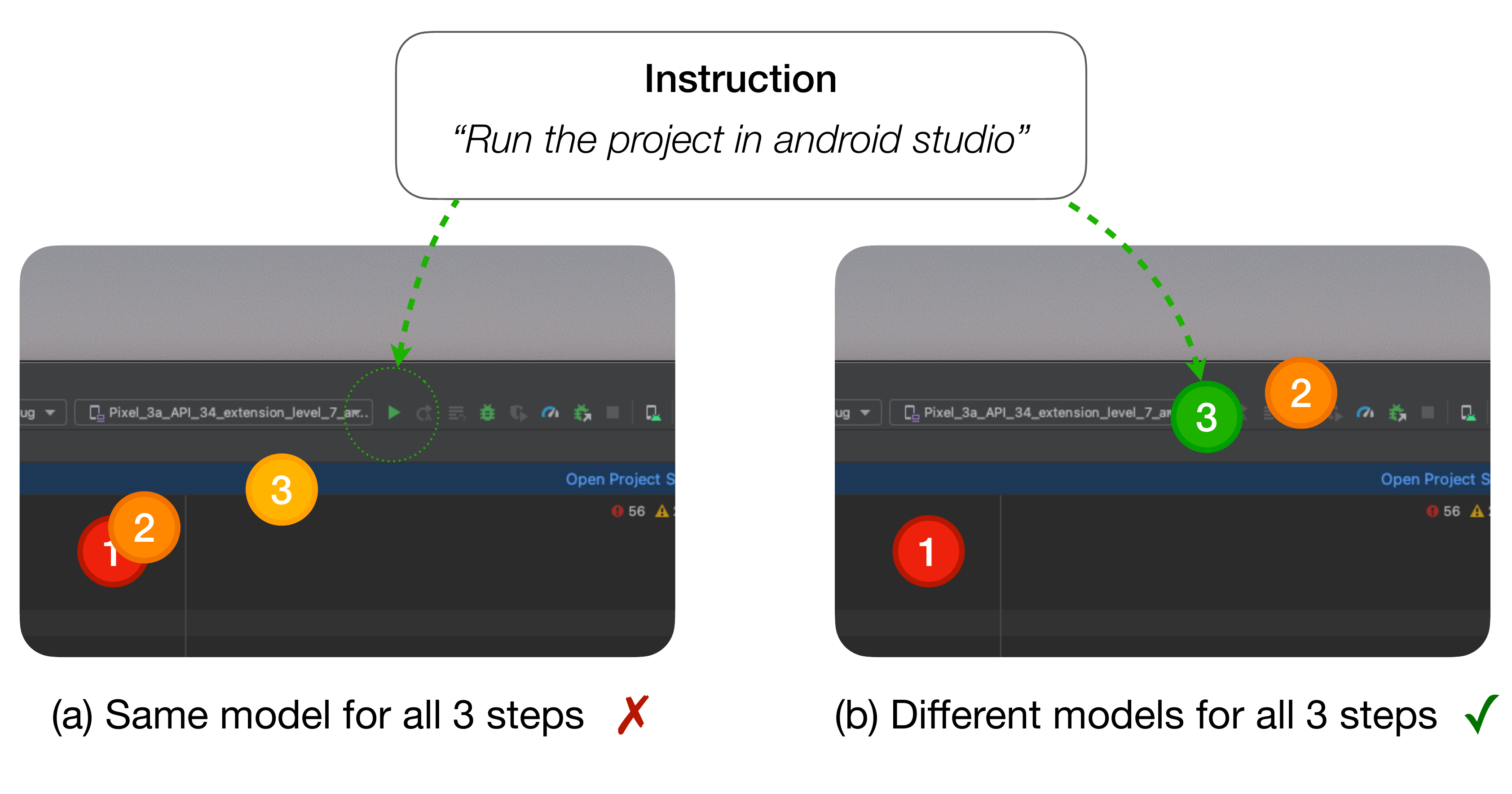}
    \caption{
    Example illustrating the effect of model combinations in the CoG framework. 
    In (a), using the same model across all three steps results in repeated errors and poor final grounding. 
    In (b), using three different models progressively refines the prediction, ultimately producing the correct result. 
    This supports our hypothesis that combining diverse models helps overcome individual blind spots and enhances final localization accuracy.
    }
    \label{fig:model_combination_example}
\end{figure}

%% file: sec/6_conclusion.tex
\section{Conclusion}
Our work demonstrates the effectiveness of a multi-step, multi-model Chain-of-Grounding (CoG) framework for complex visual understanding. By combining models with diverse capabilities, such as the proposed UI-TARS-1.5-7B $\rightarrow$ Qwen3-VL-235B $\rightarrow$ Qwen3-VL-32B \cite{qin2025uitarspioneeringautomatedgui,qwen2025qwen3vl} pipeline, we significantly enhance performance, achieving state-of-the-art results on ScreenSpot-Pro \cite{li2025screenspotpro} with 68.4\% average accuracy. This design exploits the complementary strengths of each model, compensating for individual weaknesses and yielding more accurate and robust outcomes. Our ablations show that performance is improved by: (1) iterative refinement, where each step builds on previous reasoning and corrects initial mistakes, with triple-step refinement outperforming dual-step, and dual-step outperforming single-step; (2) integrating multi-modal feedback, where combining text and visual cues provides richer grounding context; and (3) strategic heterogeneous model combinations, which leverage diverse model strengths to achieve results unattainable by any single model. Overall, our findings highlight CoG as a versatile framework for building reliable and accurate visual grounding systems and for designing an intelligent system that could understand the real-world scenario more accurately and safely.
\vspace*{-1em}
\paragraph{Limitation}
Despite these gains, several limitations remain. First, the performance of CoG is ultimately bounded by the capabilities of its base models; no guidance strategy can overcome their inherent limits. Second, compared to single-pass approaches, our iterative method incurs higher inference time and computational cost. While the framework emulates aspects of human thought processes, this does not establish that the neural network is actually “reasoning.” Third, CoG does not guarantee correct reasoning paths, and both correct and incorrect answers can arise from flawed intermediate steps. Finally, our experiments are conducted on a specific benchmark, and the generalization of our model combinations and feedback strategies to other domains remains an open question.
\vspace*{-1em}
\paragraph{Social Impacts}
The proposed framework has several potential social implications. On the positive side, more accurate visual understanding systems can support powerful accessibility tools for visually impaired users and improve user experience by making interfaces more intuitive and responsive. However, increasing reliance on complex black-box models complicates the analysis of failure cases and may lead to unintended consequences. There are also privacy concerns associated with large-scale visual data processing. Moreover, the substantial computational resources required to train and deploy such systems risk widening the digital divide, concentrating access to these capabilities among well-resourced institutions.